\documentclass[runningheads]{llncs}
\usepackage[T1]{fontenc}
 
\usepackage{accv}

\usepackage{accvabbrv}

\usepackage{graphicx}
\usepackage{booktabs}

\usepackage{float}
\usepackage{multirow}
\usepackage{wrapfig}
\usepackage{booktabs} %
\usepackage{pifont}
\usepackage{arydshln}
\usepackage{listings}
\newcommand{\xmark}{\ding{55}}
\usepackage[accsupp]{axessibility}  %

\usepackage{hyperref}

\usepackage{orcidlink}
\newcommand{\paragraphcustom}[1]{\noindent\textbf{#1}}

\begin{document}

\title{NewMove: Customizing text-to-video models with novel motions} 

\titlerunning{NewMove}

\author{
    Joanna Materzy\'nska\thanks{Work done in part as an intern with Adobe Research.}\inst{1,2} \and 
    Josef Sivic\inst{2,3} \and 
    Eli Shechtman\inst{2} \and 
    Antonio Torralba\inst{1} \and 
    Richard Zhang\inst{2} \and 
    Bryan Russell\inst{2}
}

\authorrunning{J. Materzy\'nska et al.}

\institute{Massachusetts Institute of Technology \and Adobe Research \and  CIIRC CTU \thanks{Czech Institute of Informatics, Robotics and Cybernetics at the
Czech Technical University in Prague.}}

\maketitle

\begin{abstract}

We introduce an approach for augmenting text-to-video generation models with novel motions, extending their capabilities beyond the motions contained in the original training data. With a few video samples demonstrating specific movements as input, our method learns and generalizes the input motion patterns for diverse, text-specified scenarios. 
Our method finetunes an existing text-to-video model to learn a novel mapping between the depicted motion in the input examples to a new unique token. 
To avoid overfitting to the new custom motion, we introduce an approach for regularization over videos. 
Leveraging the motion priors in a pretrained model, our method can learn a generalized motion pattern, that can be invoked with novel videos featuring multiple people doing the custom motion, or using the motion in combination with other motions.
To validate our method, we quantitatively evaluate the learned custom motion and perform a systematic ablation study. We show that our method significantly outperforms prior appearance-based customization approaches when extended to the motion customization task. Project webpage: \textit{https://joaanna.github.io/customizing\_motion/}.
\end{abstract}

\section{Introduction}
\label{sec:intro}
\begin{figure}[t!]
    \centering\includegraphics[width=\linewidth]{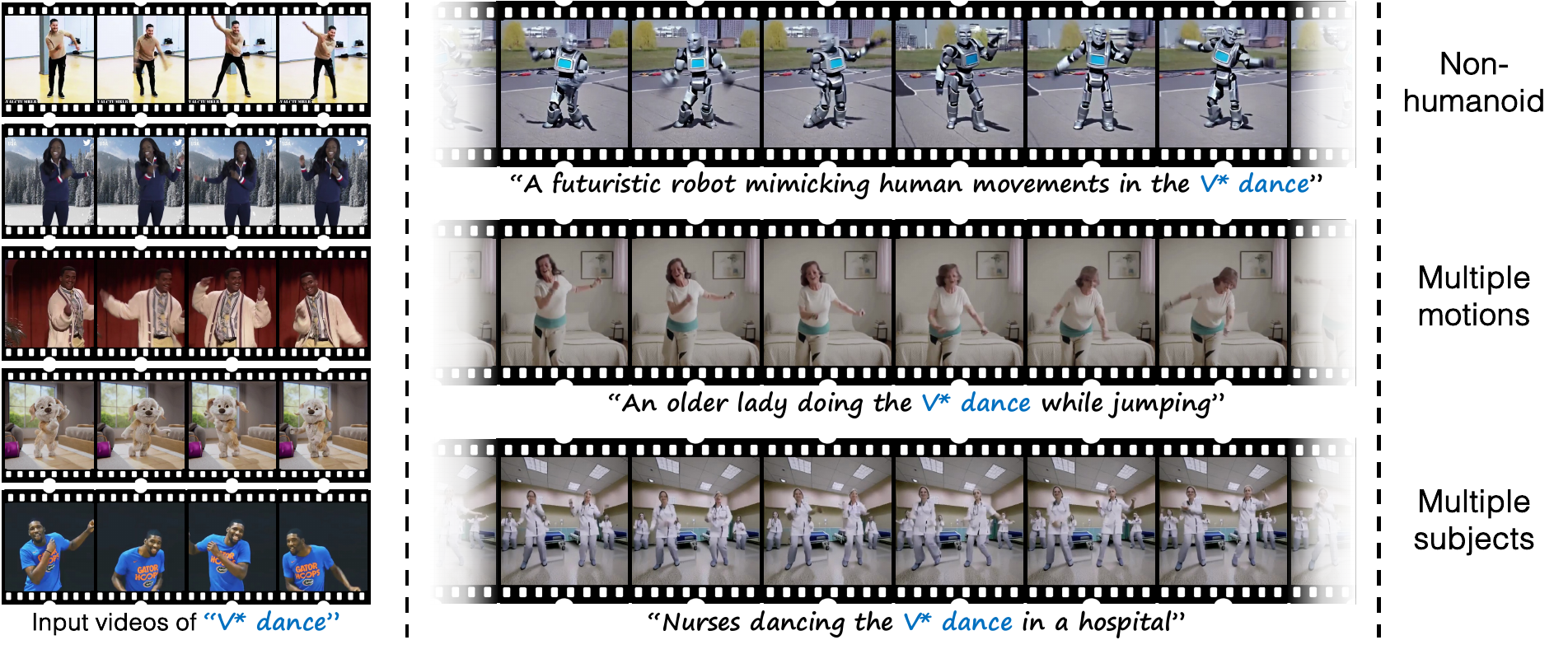}%
        \captionof{figure}{{\footnotesize \textbf{(Left)} Given a few examples (``Carlton dance''), our customization method learns the dynamic motion pattern common to the input examples and incorporates it into a pre-trained text-to-video diffusion model using a new motion identifier (``\textcolor{blue}{V* dance}''). \textbf{(Right)} Our approach, \textit{NewMove},  abstracts the motion pattern from the appearance in the input videos and enables generation of the depicted motion across a variety of novel contexts, including with a non-humanoid subject (robot, top row), multiple motions (lady, middle row), and multiple subjects (group of nurses, bottom row). {\bf To best view the results, please view our \href{https://joaanna.github.io/customizing_motion/}{website}
.}}}%
    \label{fig:teaser}
\end{figure}

Recent advancements in text-to-video synthesis have significantly pushed the boundaries of video generation \cite{blattmann2023align, wang2023videofactory, villegas2022phenaki, ho2022imagen, wang2023modelscope, zhou2022magicvideo, molad2023dreamix}.
As synthesized videos become more realistic, there is an increasing demand to provide precise control over the output, meeting specific user needs and creative visions. Consider, for example, a creator aiming to depict characters doing the `Carlton Dance' from ``The Fresh Prince of Bel-Air'' in a synthetic video with their own background and setting, as shown in Fig \ref{fig:teaser}.
If the pre-trained model does not know the specific dance, then capturing the subtle nuances of the dance's swinging arms and the precise timing of its poses through natural language is challenging, necessitating a method to instead learn from visual examples. However, the challenge is not to merely replicate the dance as in traditional motion transfer methods \cite{chan2019everybody, siarohin2019animating, siarohin2019first}, but to teach the model about the motion from a few examples (`customization') and to apply this learned motion seamlessly across diverse scenes, without relying on an original driving video.
Rather than copying a motion from a single video, we wish to capture and learn the motion's  attributes while abstracting away the appearance and context of the example setting. 
These challenges lead us to a question: How can we leverage a pre-trained text-to-video model's priors about motion and appearance to augment them with new motions, and subsequently generate these new motions in novel settings?

Traditional methods in video generation have offered detailed control through techniques like conditional generation \cite{yang2018pose, wang2018video, esser2023structure} and motion transfer \cite{gleicher1998retargetting, chan2019everybody, siarohin2019animating, singer2022make,wiles2018x2face}, enabling the precise replication of specific motions depicted in the input video. However, these methods are limited by their reliance on the original video's scene structure and geometry, effectively locking the motion to the context of the driving video. Such an approach restricts the ability to apply the motion more broadly, as it does not truly learn and abstract the motion pattern for flexible use across varied scenes and subjects. Developing methods that can generalize motion patterns from examples for use in diverse settings presents a significant challenge, as it requires methods that go beyond simple replication to allow for the motion's adaptation to new environments. 
A more similar scenario to ours focuses on customization of subjects in text-to-image models \cite{ruiz2023dreambooth, kumari2023multi, gal2022image, chen2023subject, gal2023encoder, wei2023elite, han2023svdiff, tewel2023keylocked}. However, unlike learning a novel dynamic motion from few example videos, these methods are restricted to still subjects.
In Tab.~\ref{tab:baselines} we compare the design of related approaches to ours, and discuss them in more detail in Section~\ref{sec:related}.

\begin{table}[t!]
\centering
\caption{Comparison of our method across different techniques for controllable video / image generation. Here, $\mathcal{G}_I$, $\mathcal{G}_{I\rightarrow V}$ and $\mathcal{G}_V$ denote text-to-image, image-to-video and text-to-video generation models, respectively. Symbols $\mathcal{I}$ and $\mathcal{V}$ denote input images and videos, $\mathcal{\hat{I}}$ and $\mathcal{\hat{V}}$ denote output images and videos, and $\tau$ denotes a text prompt. Superscripts refer to {\color{violet} learned or transferred concepts} , either appearance${^\dagger}$ or motion${^\circ}$.
Our method is the only one that transforms a text-to-video motion prior (\textcolor{red}{$\mathcal{G}_V$})
to a model with customized motions (\textcolor{blue}{$\mathcal{G}^{\circ}_V$}, \textcolor{blue}{$\mathcal{G}^{\circ\dagger}_V$}) such that the learned concept ({\color{violet} $\tau^{\circ}, \tau^{\circ\dagger}$}) can be invoked at the test time.}
\label{tab:baselines}

\resizebox{0.95\textwidth}{!}{%
\begin{tabular}{lcc}
\toprule
Methods & Customization & Application \\
\midrule

Image animation \cite{chen2023videocrafter1} & None & $\mathcal{G}_{I\rightarrow V}(\mathcal{I})\rightarrow \mathcal{\hat{V}}$ \\
Text-to-video generation \cite{chen2024videocrafter2, blattmann2023align, wang2023videofactory, villegas2022phenaki, ho2022imagen, wang2023modelscope, zhou2022magicvideo, molad2023dreamix} & None & $\textcolor{black}{\mathcal{G}_V}(\tau)\rightarrow \mathcal{\hat{V}}$ \\
Structure-conditioned video generation \cite{esser2023structure} & None & $\textcolor{black}{\mathcal{G}_V}(\tau, V)\rightarrow\mathcal{\hat{V}}$ \\
Zero-shot text-driven motion transfer \cite{geyer2023tokenflow, yatim2023space} & None & $ \mathcal{V},  \textcolor{black}{\mathcal{G}_{V}}({\color{violet}{\color{violet}\tau^{\circ}}}) \rightarrow \mathcal{\hat{V}} \text{ or } \mathcal{V}, \mathcal{G}_{I}({\color{violet}\tau^{\circ}})  \rightarrow \mathcal{\hat{V}}$ \\ %

Image customization \cite{gal2022image, ruiz2023dreambooth, kumari2023multi} &  $\mathcal{G}_I\xrightarrow{\mathcal{I}} \mathcal{G}^{\dagger}_I $ & $\mathcal{G}^{\dagger}_I({\color{violet}\tau^{\dagger}})\rightarrow \mathcal{\hat{I}}$  \\

Animating text-to-image models (AnimateDiff v1) \cite{guo2023animatediffv1} & $\mathcal{G}_I^{\dagger}, \textcolor{red}{\mathcal{G}_V}\rightarrow \mathcal{G}^{\dagger}_V$  &  $\mathcal{G}^{\dagger}_V(\tau)\rightarrow \mathcal{\hat{V}}$ \\

One-shot video tuning \cite{wu2023tune} & $\mathcal{G}_I\xrightarrow{\mathcal{V},\tau}\textcolor{blue}{\mathcal{G}^{\circ}_V}$ & $\textcolor{blue}{\mathcal{G}^{\circ}_V}({\color{violet}\tau^{\circ}})\rightarrow \mathcal{\hat{V}}$ \\ 
\cdashline{1-3}
\vspace {-3mm} \\
Ours (motion customization) & $\textcolor{red}{\mathcal{G}_V}\xrightarrow{\mathcal{V}} \textcolor{blue}{\mathcal{G}^{\circ }_V}$ & $\textcolor{blue}{\mathcal{G}^{\circ }_V}({\color{violet}\tau^{\circ}})\rightarrow \mathcal{\hat{V}}$ \\

Ours (motion and appearance cust.; see \href{https://joaanna.github.io/customizing_motion/}{our website}) & $\textcolor{red}{\mathcal{G}_V}\xrightarrow{\mathcal{V},\mathcal{I}} \textcolor{blue}{\mathcal{G}^{\circ \dagger}_V}$ & $\textcolor{blue}{\mathcal{G}^{\circ \dagger}_V}({\color{violet}\tau^{\circ \dagger}})\rightarrow \mathcal{\hat{V}}  $ 

\\
\bottomrule
\end{tabular}
}
\end{table}

Incorporating new motions in text-to-video models ({\em motion customization}) faces unique challenges. The first issue is determining which model parameters to adjust for capturing distinct motions from few examples, balancing the need for model flexibility against the risk of overfitting due to limited data. Another challenge is maintaining the model's existing knowledge without having new customizations overshadow or alter learned concepts, such as causing a bias towards a specific dance like the Carlton. Additionally, it is critical to separate motion from appearance so that learned motions can be applied to new scenes. 

In this work, we address these challenges and introduce a novel method for adapting text-to-video diffusion models to recognize and generate new motions,
such as local motions (human movement) or global (camera motions), independent of the performer's appearance or scene's appearance, \ie, customizing text-to-video models with novel motions. By training the model with a few example videos of a new motion, we assign this motion to a unique text token (``V*''). This assignment allows us to produce videos featuring various subjects performing the new motion, as illustrated in Fig \ref{fig:teaser}. 
Through an ablation study, we explore customizing various model parameters, highlighting that adjustments to temporal layers and spatial cross-attention mechanisms strike an optimal balance between model adaptability and the risk of over-fitting. 
To preserve the model's original motion knowledge and avoid neglecting existing motion concepts, we develop a unique regularization strategy specifically for video content.
To prioritize learning about motion over appearance, we introduce a new sampling strategy that focuses on samples that determine  coarser visual structure. %

To evaluate our method we perform two qualitative studies. The first is a user study comparing our method against two baselines across diverse scenarios, including videos with multiple people and varying subject sizes, where the custom motions are sourced from internet videos. The second evaluation employs gestures from the Jester dataset, aiming to systematically examine our method's effectiveness in internalizing and recognizing motions via an off-the-shelf classifier. {\bf Our contributions are}: (i) introducing an approach that leverages the rich motion and appearance prior information contained in a pretrained text-to-video generation to augment it with novel motions, (ii) demonstrating the generalization abilities of our method to diverse scenes and multiple subjects, and (iii) lastly a qualitative and quantitative evaluation of our method.

\section{Related Work}
\label{sec:related}
\paragraphcustom{Motion transfer.}
Motion transfer is a well established task \cite{gleicher1998retargetting, chan2019everybody, siarohin2019animating, singer2022make,wiles2018x2face}. The task involves capturing motion from one video and applying it to the target subject given by an image or a video. These methods explicitly model the pose of the source motion and focus on modelling humans. More recently, the task of text-driven motion transfer has been introduced \cite{yatim2023space, geyer2023tokenflow}. Originally, Geyer \etal \cite{geyer2023tokenflow} leveraged a pre-trained text-to-image model to generate videos that preserve the overall motion of a source video and provide some flexibility with respect to the dynamic target subject, without tuning the original model. Yatim \etal \cite{yatim2023space} proposed a similar approach that instead utilizes a text-to-video model. 
Other approaches \cite{esser2023structure, chai2023stablevideo, ceylan2023pix2video}, ensure output videos adhere to reference videos by guiding the generation with depth information.
Our approach differs from these prior works in that we learn a generalized motion pattern from a few video examples. That is, instead of directly copying the motion or the overall video motion pattern, our method allows for invoking the motion in vastly different text-defined scenarios. This setting includes creating videos where multiple individuals execute the motion simultaneously, or blending this motion with other distinct movements. 
\begin{figure*}[t]
\centering\includegraphics[width=\linewidth]{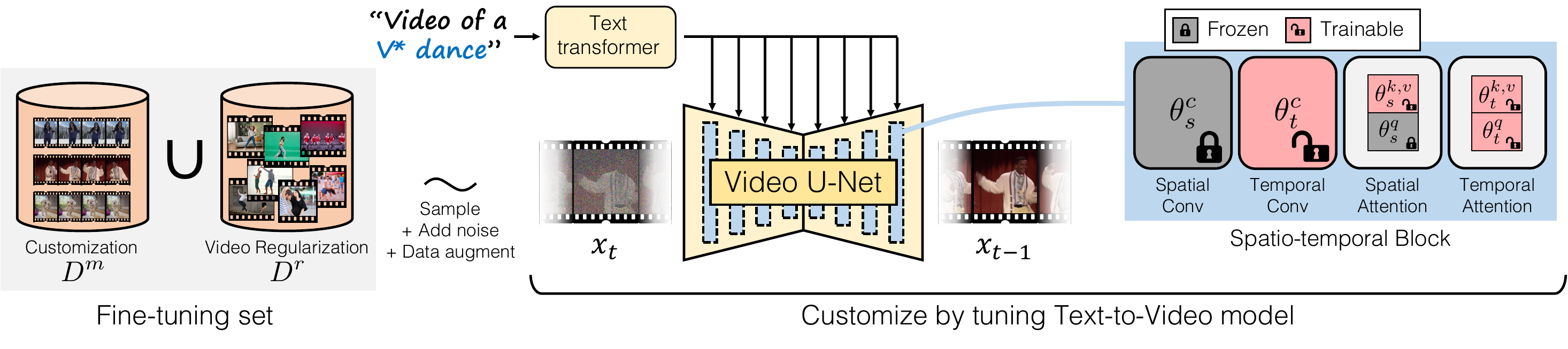}%
    \caption{\textbf{Overview.} Given a small set of exemplar videos, our approach fine-tunes the U-Net of a text-to-video model using a reconstruction objective.
    The motion is identified with a unique motion identifier and can be used at test time to synthesize novel subjects performing the motion. To represent the added motion but preserve information from the pretrained model, we tune a subset of weights -- the temporal convolution and attention layers, in addition to the key \& value layers in the spatial attention layer.  A set of related videos is used to regularize the tuning process.}
    \label{fig:methodfig}
\end{figure*}
\paragraphcustom{Video synthesis with image generation models.}
Previous works repurposed text-to-image generation models for video generation. Wu \etal \cite{wu2023tune} used a pretrained image-to-video model and a source video to stylize the video with a text condition. Similarly, several methods \cite{chen2023controlavideo, zhang2023controlvideo, khachatryan2023text2video} can generate a video conditioned on control signals such as depth maps of edges using a pre-trained text-to-image model. Guo \etal \cite{guo2023animatediffv1} learn a motion prior network with text-video data that can be used together with personalized text-to-image models to generate animated images. Our work differs in that we use a video generation model directly to learn new motions and, rather than animating images, our method produces videos with complex dynamic motions.  \paragraphcustom{Text-to-video models.}
Recently, several methods have demonstrated text-to-video generation \cite{blattmann2023align, wang2023videofactory, villegas2022phenaki, ho2022imagen, wang2023modelscope, zhou2022magicvideo, molad2023dreamix}. 
An important line of research involves studying architectures and optimization for training video generation models. A high-quality model is a prerequisite for our method. In this work, we use the model from Wang \etal \cite{wang2023modelscope}. Molad \etal \cite{molad2023dreamix} show qualitative results of adding a custom subject to the base video model. They, however, do not explore customization beyond static objects \cite{molad2023dreamix}. The idea of multi-modal customization, however, facilitates user-controllability and creative applications, since no matter how powerful the model, there are always novel subjects and motions that cannot be all contained in the training set used to train the base video generation model.\paragraphcustom{Text-to-image and text-to-video customization.}
Text-to-image customization is a related task to ours, where a text-to-image diffusion model is augmented with a custom object \cite{ruiz2023dreambooth, kumari2023multi, gal2022image, chen2023subject, gal2023encoder, wei2023elite, han2023svdiff, tewel2023keylocked}. Gal \etal \cite{gal2022image} showed that this mapping can be created purely by optimizing a text embedding of the textual description. Kumari \etal \cite{kumari2023multi} proposes a method that fine-tunes selected weight matrices in the generator network.
More recent approaches investigate efficient formulations for modifying an approximation of the parameters responsible for the text-image mapping in the generation network \cite{han2023svdiff, tewel2023keylocked}. Other methods, alternatively to adjusting the model's parameters, introduce an image encoder that can be directly plugged into the pre-trained network \cite{gal2023encoder, wei2023elite, zhou2023enhancing}. Alternatively to customizing individual concepts, Huang \etal \cite{huang2023reversion} propose to learn relations between two subjects from a set of images via a similar technique to Gal \etal \cite{gal2022image}.
Similarly, image animation techniques use a video motion prior to turn still images into videos \cite{guo2023animatediffv1, chen2023videocrafter1}.
The text-to-image customization is limited to static objects due to the static nature of still images. In contrast, we focus on customization with respect to a given motion depicted in a small set of videos, where the task is to learn the dynamic motion pattern.  
More recently, several concurrent and yet unpublished works ~\cite{zhao2023motiondirector, wei2023dreamvideo, jeong2023vmc, zhang2023motioncrafter, wu2023lamp, guo2023animatediffv2} highlight the growing interest in motion customization, demonstrating its importance in current research. These works explore various facets of the problem, including transferring subject motion from a video~\cite{jeong2023vmc,zhang2023motioncrafter}, customizing appearance and motion from images and videos~\cite{wei2023dreamvideo}, animating high-resolution images~\cite{wu2023lamp}, animating images with custom camera motion patterns ~\cite{guo2023animatediffv2}, and separating motion from appearance~\cite{zhao2023motiondirector}. While these works are exciting, our approach uniquely stands out in the context of this recent work by abstracting dynamic motion from multiple videos, enabling the creation of new, diverse videos that showcase different scenes, combine different motions, and depict various subjects.

\section{Approach}
\label{sec:method}

\subsection{Text-to-Video Diffusion Model Preliminaries}
\label{sec:preliminaries}

Diffusion models are probabilistic models that can approximate distributions through a gradual denoising process \cite{ho2020denoising, song2020denoising}. 
Given a Gaussian noise sample, the model learns to iteratively remove a portion of noise until a sample is within the approximated distribution by minimizing the $L_2$ distance of the predicted noise and the sampled noise.
Latent Diffusion Models (LDMs) operate in latent rather than pixel space, encoding videos into a lower-dimensional vector via an encoder-decoder model to simplify the denoising process \cite{rombach2021high}.

Specifically, a video $x$ is represented with a latent vector, and a text condition $c$ is encoded through a text-embedding from a pre-trained text encoder. The initial noise map $\epsilon$ is sampled from a Gaussian distribution $\epsilon \sim \mathcal{N}(0, 1)$. 
For the diffusion timestep $t$ sampled from a probability density $t \sim f(t)$, 
the noisy latent sample can be computed from the noise prediction by $x_t = \sqrt{\alpha_t} x_0+\sqrt{1-\alpha_t} \epsilon$, where $x_0$ is the original video and $\alpha_t$ controls the amount of noise added at each diffusion timestep according to a noise scheduler \cite{song2020denoising, ho2020denoising}.  The model $\epsilon_\theta$ with parameters $\theta$ is trained with the following weighted denoising loss,
\begin{equation}
    \mathcal{L}_\theta(x, c) = \mathop{\mathbb{E}}_{\substack{\epsilon\sim \mathcal{N}(0, 1) \\ t\sim f(t)}} [ w_t\|\epsilon_\theta(x, \epsilon, c, t) - \epsilon  \|^2_2],
\label{eqn:opti}
\end{equation}
where $w_t$ is a user-defined variable that controls the sample quality.
At inference time, a new video can be generated by sampling a Gaussian noise sample $\epsilon$ and a text prompt, denoising them with the learned model.

\subsection{Approach for Motion Customization}
\label{sec:motion_customization}
We illustrate our overall approach for motion customization in Figure~\ref{fig:methodfig}. 
Let the motion be represented through a small exemplar set of videos and corresponding text prompts $D^m = \{(x^{m}, c^m)\}$.
The motion can be performed by different subjects across different backgrounds, and the commonality within the videos is purely the dynamic movement.
We choose a generic textual description across the videos, such as ``a person doing the V*''. In practice, we select rare tokens like ``pll'' as ``V*''. 
To customize a text-to-video model's parameters $\theta$,
we fine-tune the model's parameters by minimizing the diffusion objective $\mathcal{L}_\theta$ summed over the exemplar set $D^m$,
\begin{equation}
\min_\theta \sum_{(x, c) \sim D^m}  \mathcal{L}_\theta(x, c).
    \label{eqn:full_loss}
\end{equation}
At test time  we can generate novel videos of the target with variations controlled by text. 
Customizing text-to-video generation models has, however, several open technical questions that need to be addressed: (i) Which model parameters to customize? (ii) How to prevent forgetting of already learnt concepts? (iii) How to disentangle motion and appearance? 
We address the above open questions with the following three technical components, as described next.

\paragraphcustom{Choice of customization parameters.}
The quality and generalization ability of the novel motion depends on the choice of the model parameters updated during customization. 
A text-to-video diffusion model $\epsilon_\theta$ has parameters $\theta$ that can be categorized to those operating on the temporal dimensions $\theta_t\subset\theta$ and those operating on the spatial dimensions (per-frame) $\theta_s\subset\theta$. 
Let $\theta_s^{k,v}\subset\theta_s$ be the keys and values parameters of the spatial cross-attention modules.
The temporal layers $\theta_t$ are transformers and temporal convolutions, and are responsible for modelling temporal patterns across frames. In Section~\ref{sec:experiments}, we empirically show that the temporal layers alone do not effectively model a new motion pattern due to time-varying appearance changes. For example, consider an object rotating in 3D, which requires the model to generate the appearance of disoccluded surfaces.  
To faithfully learn a motion pattern, we also modify a subset of parameters in the spatial layers of the model. 
As illustrated in Figure~\ref{fig:methodfig}, our approach fine tunes the spatial keys/values $\theta_s^{k,v}$ and temporal $\theta_t$ parameters. 
Note that in image customization~\cite{kumari2023multi, tewel2023keylocked} and model editing~\cite{gandikota2023unified}, it has been shown that the spatial keys and values of the cross-attention modules in text-to-image models are sufficiently powerful to represent the appearance of new concepts. We show the effectiveness of tuning these parameters for motion customization.

\paragraphcustom{Preventing forgetting via video regularization.}
Prior work has shown that directly optimizing Equation (\ref{eqn:full_loss}) leads to  forgetting related concepts or the concept category \cite{ruiz2023dreambooth, lu2020countering, lee2019countering}. For example, if the concept is a specific person, all people start resembling that person. To mitigate this issue, prior work has utilized a regularization set that ensures that this knowledge is preserved. 
For example, Ruiz \etal \cite{ruiz2023dreambooth} proposed collecting a regularization set via generation from the original model. 
Kumari \etal \cite{kumari2023multi} proposed using pairs of real images and text. 

In contrast to prior work, we seek to mitigate forgetting of related motions that the model has already learned. To address this goal, we consider a video-based regularization. 
Let $D^r$ be a regularization set of paired videos with text descriptions that have similar but not identical motion to the target custom motion videos $D_m$. For example, when learning a gesture from the Jester dataset, we chose a regularization set of real videos containing people sitting in front of a webcam. Since this type of video data might be difficult to come across, using a regularization set of real videos in-distribution to the training set is also helpful. For customization, we optimize the diffusion objective $\mathcal{L}_\theta$ over the target custom videos $D^m$ and regularization set $D^r$: 
\begin{equation}
    \min_\theta \sum_{(x, c) \sim D^m \cup D^r } \mathcal{L}_\theta(x, c).
\label{eqn:reg}
\end{equation}
Empirically, we show in Section~\ref{sec:experiments} that using real videos for the regularization set is superior to using generated videos from the model. We find that using generated videos degrades the quality of the customized model substantially.

\paragraphcustom{Disentangling motion from appearance.}
To facilitate learning the common motion in $D^m$, during the training process, we aim to put emphasis on the motion pattern of the videos, rather than the appearance or subject presenting the motion. For example, in the case of the `Carlton' dance in Figure \ref{fig:teaser} we wish to capture the motion pattern of the dance rather than the appearance of the background or the individual performers. The denoising process in diffusion models samples a Gaussian noise and then gradually removes the noise. The initial noise as well as early denoising steps have large influence on the output overall dynamic structure of the video, whereas the later stages correspond to finer details \cite{wang2023diffusion}. To focus on the dynamic structure of the videos and de-emphasize the appearance of the subjects performing the motion, we define a timestep sampling strategy. In particular, we build on~\cite{huang2023reversion} who develop a non-uniform sampling strategy for generating still images. Here we adopt it for generating dynamic video content. 
In detail, instead of uniformly sampling the denoising steps in Equation (\ref{eqn:opti}), we define a probability distribution $f_{\alpha}(t) = \frac{1}{T}(1-\alpha \cos(\frac{\pi t}{T}))$  over the timesteps that focuses more on the earlier denoising steps and hence emphasizes learning the overall dynamic motion pattern rather than the fine appearance details of the background or the individual subjects. The $\alpha$ parameter increases the skewness of the function. We use $\alpha = 0.5 $ for all of our experiments.

\section{Experiments}
\label{sec:experiments}

\subsection{Qualitative Comparison and a User Study}
\label{sec:qualitative}
Compared to the published prior work, as shown in Tab.~\ref{tab:baselines}, 
ours is the only method that transforms a text-to-video motion model (\textcolor{red}{$\mathcal{G}_V$}) to a model with customized motion (\textcolor{blue}{$\mathcal{G}^{\circ}_V$}) or customized motion and appearance (\textcolor{blue}{$\mathcal{G}^{\circ\dagger}_V$}) (see the \href{https://joaanna.github.io/customizing_motion/}{website} for our joint motion and appearance customization). 
We adapt prior work that learns or transfers concepts (${\color{violet} \tau^{\circ}}$) to compare with our problem setting.
In particular, we focus on   ``Image customization'' methods that customize the appearance of the subject and reference-based (``One-shot video tuning'' or ``Zero-shot text-driven motion transfer'') methods that transfer motion from a single input source video. We adapt these methods to closer match our problem definition as described next. We show results for customizing the base model to selected human gestures, which the base model does not know. For this purpose we use videos from the Jester Dataset~\cite{materzynska2019jester}, which contains crowd-sourced videos of diverse actors performing a gesture in front of a static camera.

\paragraphcustom{Comparison with image customization methods.}
We select three well-established image customization approaches: Textual Inversion \cite{gal2022image},  Dreambooth \cite{ruiz2023dreambooth} and Custom Diffusion \cite{kumari2023multi}, and adapt them to the motion customization task. Textual Inversion \cite{gal2022image} optimizes a text token embedding to learn a novel subject. For Dreambooth \cite{ruiz2023dreambooth}, we optimize all spatial layers of the model and use a synthetic video regularization set following their prior preservation loss. Custom Diffusion trains the key and value projection matrices in the cross-attention layers of text-to-image diffusion models, optimizes a text token embedding, and uses a regularization set with real videos.

\begin{figure}[t!]
    \centering
    \includegraphics[width=0.9\textwidth]{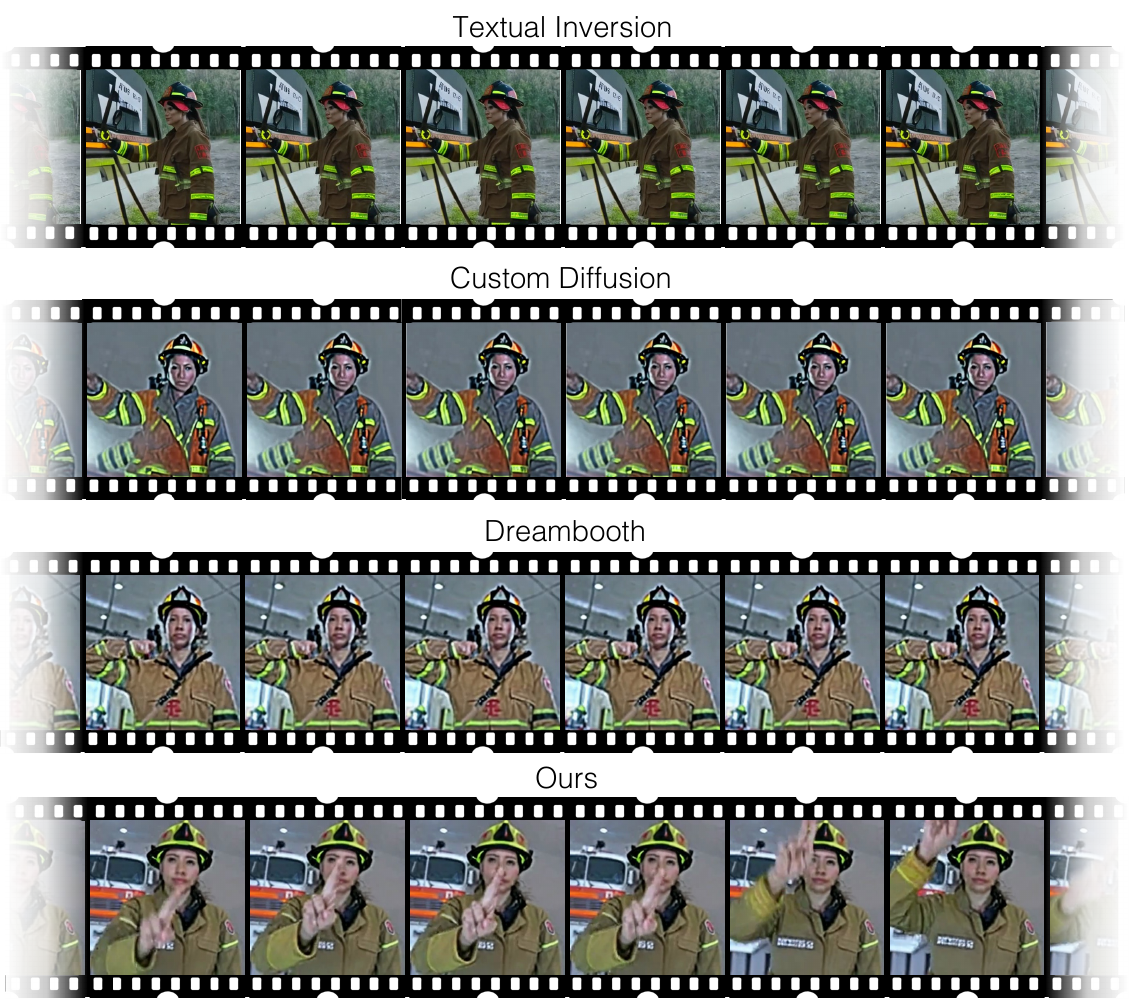}
    \caption{{\bf Visual comparison with baseline methods.} Examples of learning a customized motion \textit{Sliding Two Fingers Up} from the Jester dataset with prompt ``{\it A female firefighter doing the \textcolor{blue}{V* sign}}''. Baseline methods (top three rows) fail to capture the motion and produce a temporally coherent video.}
    \label{fig:sota}
\end{figure}

We choose at random 10 videos from the validation set from the Jester dataset\cite{materzynska2019jester} class ``Sliding Two Fingers Up'' as our training set. To learn the motion, we caption the videos with a generic text description ``a person doing a V* sign''. For the motion learning regularization set, we choose 50 videos from each of the dataset classes $\{$\textit{Doing Other Things}, \textit{No gesture}$\}$ and assign them text captions ``a person doing other things'' and ``a person in front of a computer''.
We generate results with a prompt ``{\it A female firefighter doing the \textcolor{blue}{V* sign}}''. Fig \ref{fig:sota} shows that, in contrast to ours, the baseline methods fail to capture the motion and produce a temporally smooth video.

\begin{figure}[t!]
    \centering
    \includegraphics[width=1\linewidth]{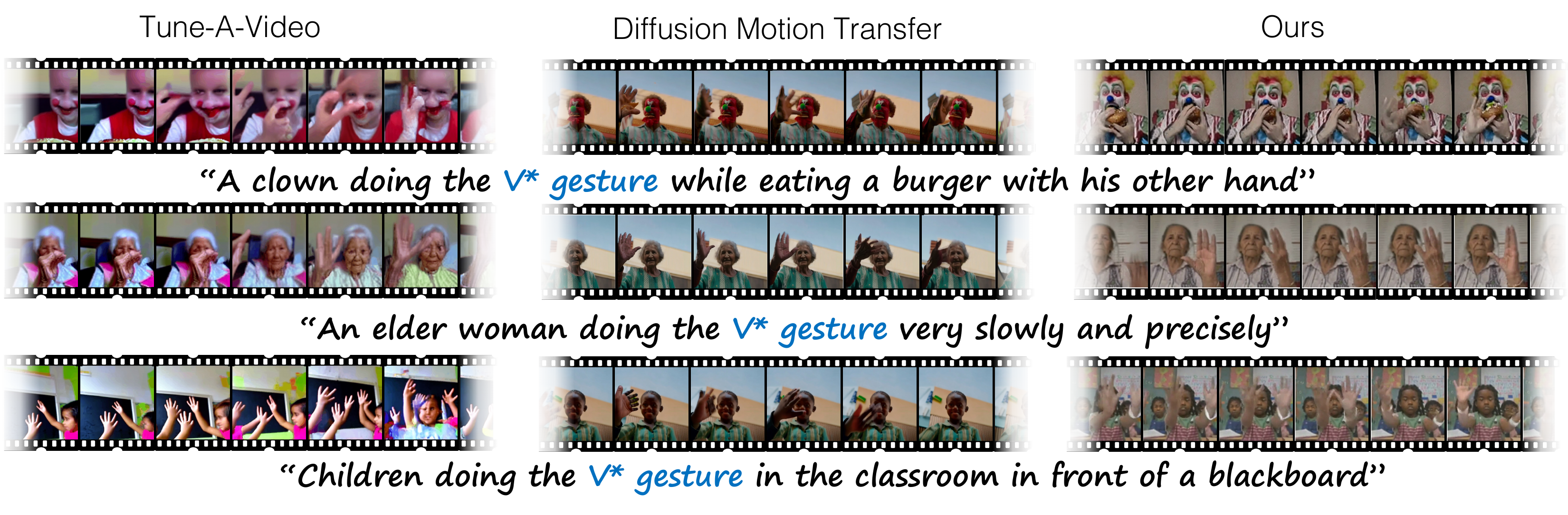}
    \caption{{\bf Text-driven motion transfer methods versus our method trained on few examples of a custom motion ``Shaking Hand''.}
Our method seamlessly renders a custom motion in novel scenarios. Despite the training videos showing only a single actor performing one motion, our method generates the custom motion alongside another action (doing the gesture while eating a burger'') and varies timing (doing the gesture slowly and precisely'') or involves multiple people (``children''). In contrast, both baselines fail to generalize or produce temporally coherent videos.
    }
    \label{fig:reference}
\end{figure}

\paragraphcustom{Comparison with a reference-based method.}
\label{sec:reference_based}
In Figure \ref{fig:reference} we compare our approach to two reference-based methods \cite{wu2023tune, yatim2023space}. We adapt Tune-A-Video to fine-tune from multiple videos using the same training set as our setup. We test the generalization abilities of the methods, firstly combining the novel motion with a different motion: ``doing the V* gesture while eating a burger with other hand''. We can see that our method can generate the two motions. Next, we test the motion variation. We show that we can manipulate the execution of the motion with a natural language description; here we specify ``doing the V* gesture very slowly and precisely''. Lastly, we show that our method can generalize to multiple people doing the motion while preserving the temporal consistency of the video. The Tune-A-Video and Diffusion Motion Transfer baseline produces qualitatively worse results and do not generalize to multiple people performing a motion, motion variations and combinations with other movements (Figure \ref{fig:reference}). 

\begin{figure}[t!]
    \centering
    \includegraphics[width=0.95\linewidth]{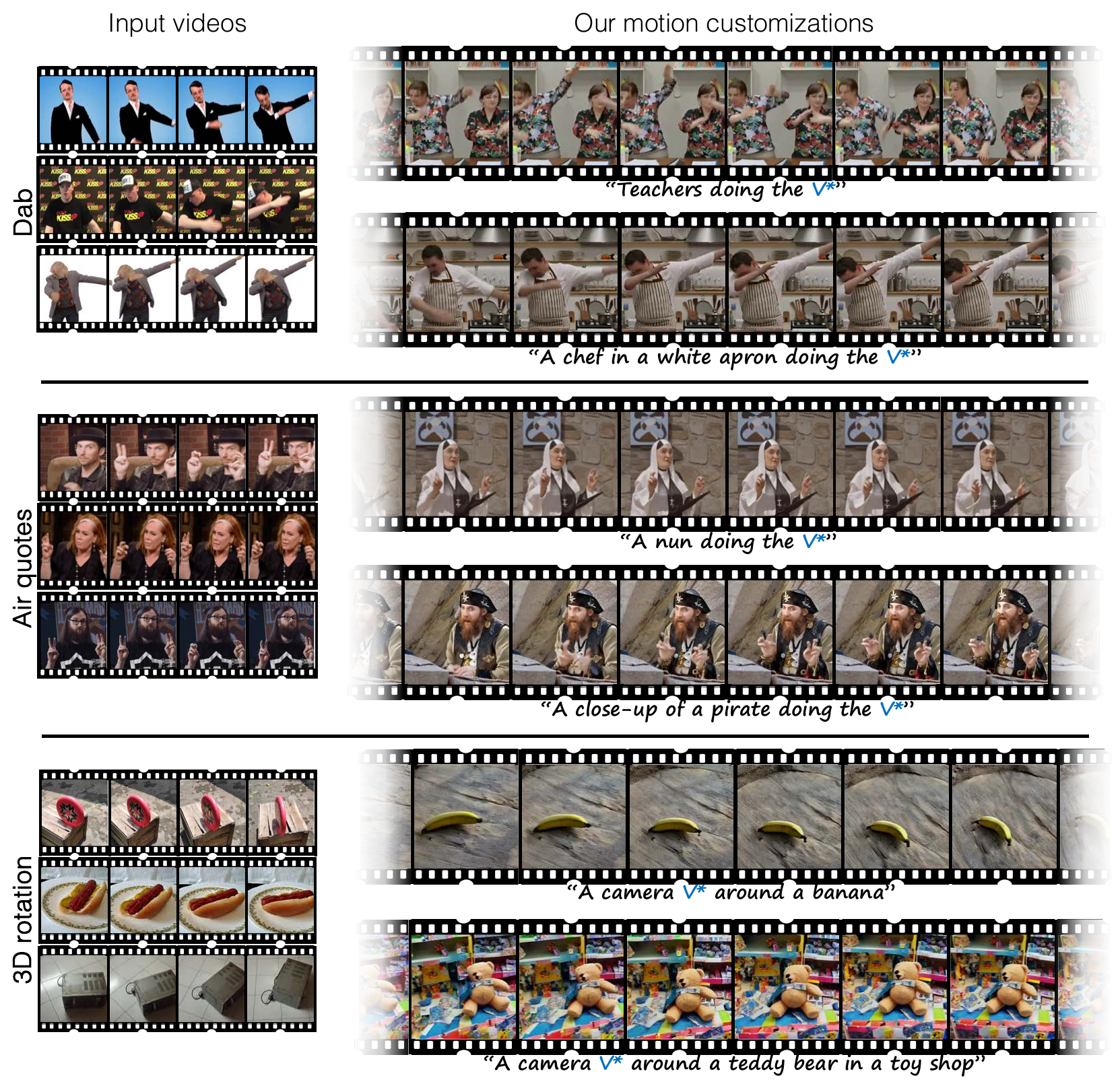}
    \caption{{\bf Qualitative results of our method.} We demonstrate two custom motions: Dab and Air quotes, trained using collected internet examples as well as a 3D camera rotation trained with examples from the CO3D dataset \cite{reizenstein2021common}.
    Our method can generalize to unseen subjects and multiple people performing the action.}
    \label{fig:qualitative}
\end{figure}

To further demonstrate the effectiveness of our approach on a broader class of motions, we choose the motions:  \textit{Carlton}, \textit{Dab}, \textit{Airquotes}, %
and \textit{3D rotation}, which the original model does not know or cannot reliably perform. For the first two %
motions, we curate videos from the internet searching for the description of the motion. 
We use between 5 and 6 videos for each motion and caption the videos with a generic text description such as ``A person doing the V* dance'', ``A person doing the V* sign'' or ``A person V*''. To learn 3D rotation, we select 10 sequences of distinct objects from the CO3D Dataset \cite{reizenstein2021common} and use as captions text prompt ``A camera V* around an [object]'', where [object] is an object~class. 
\begin{wrapfigure}[6]{r}{0.5\linewidth}
\setlength{\intextsep}{1pt}
\centering %
\includegraphics[width=1\linewidth]{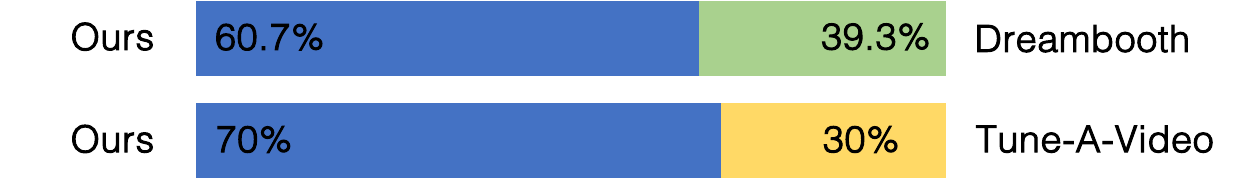}
\caption{User preference comparison.}
\label{tab:user_study}
\end{wrapfigure}
Due to the lack of readily available regularization set of similar motions, we select 100 video-text pairs from the WebVid-10M Dataset \cite{Bain21}. The results are shown in Figure \ref{fig:qualitative}. For visualization, we query for objects not present in the training videos.

\paragraphcustom{User Study.} To further demonstrate the benefits of our approach, 
we conducted a binary forced-choice user study on a diverse set of dynamic scenes involving complex motions. Involving 6 users, we assessed 50 prompts per motion (Carlton, Airquotes, and Dab, see Figures \ref{fig:teaser}, \ref{fig:qualitative}), resulting in 300 votes. We presented users with videos from our method and a baseline (DreamBooth and Tune-A-Video), presented in a random order, and asked users to select the video that best matched the input motion. Users select our method more often (Figure \ref{tab:user_study}), with statistical significance (p-values<0.01), further confirming the automatic quantitative evaluation performed in Section \ref{sec:quantitative}. 

\subsection{Automated quantitative evaluation via recognition}
\label{sec:quantitative}
In this section we introduce an automated quantitative evaluation of the generated videos by employing pre-trained descriptors \cite{radford2021learning, pizzi2022self}, and motion-specific classifiers \cite{zhou2017temporalrelation}. First, we introduce the used dataset and describe the proposed metrics. This automatic evaluation protocol then allows us to perform an exhaustive ablation study of the different components of our method. Finally, we use this automatic evaluation to compare our approach to other competing methods. 

\paragraphcustom{Dataset.}
Evaluating the motion pattern separately from the appearance is challenging. We evaluate on the Jester dataset \cite{materzynska2019jester}, which requires gesture recognition models to generalize beyond appearance and learn the specific motion pattern.
The dataset contains 148,092 crowd-sourced videos of 27 human gestures by diverse actors performing a gesture in front of a static camera. We select 5 gesture classes as our target motions -- $\{$\textit{Swiping Down}, \textit{Drumming Finders}, \textit{Rolling Hand Forward}, \textit{Shaking Hand}, \textit{Sliding Two Fingers Up}$\}$. From each class, we choose at random 10 videos from the validation set as our training set. To learn the motion, we caption the videos with a generic text description ``a person doing a V* sign''. As a regularization set for the motion learning, we also choose 50 videos from each of the dataset classes $\{$\textit{Doing Other Things}, \textit{No gesture}$\}$ and assign them text captions ``a person doing other things'' and ``a person in front of a computer''. We design a test set containing 100 text prompts across 3 random seeds, that detail a person's appearance, for example ``A female firefighter doing the V* gesture'' or ``A Japanese toddler doing the V gesture''. 
\paragraphcustom{Motion accuracy score.}
The Jester dataset has been widely adopted for the task of action recognition. We leverage this for our motion accuracy score, measuring gesture recognition accuracy with a video classifier, pre-trained on the Jester training set \cite{zhou2017temporalrelation}.The classifier reaches 94.78\% accuracy for all gesture classes, fitting our evaluation needs. Test set prompts often yield videos matching the classifier's training layout with a single person, thus being in-distribution.
\paragraphcustom{Text alignment scores.}
To regulate overfitting, we measure how faithfully the model adheres to the appearance described in the text prompt with an appearance score defined as the mean CLIP score between the generated frames and the part of the text prompt describing the person's appearance (\eg, ``A female firefighter''). We empirically find that the {\bf appearance score} is a good proxy to determine when the model starts to overfit. For a fair comparison across all models, we fine-tune each model until this appearance score reaches a chosen value (here 0.265), determined empirically based on the visual quality of the results. To assess how well the generated video corresponds to the overall text prompt (both motion and appearance), we measure the {\bf text alignment score} as a mean CLIP score between the video frames and a full caption written in natural language. For example, for the class \textit{Swiping Down} and a text prompt  ``A female firefighter doing the V* gesture'', a full caption would be ``A female firefighter swiping hand down.'' We use this score as another criterion to compare different methods.
\paragraphcustom{Copying score.} We report a copying score {\bf  (``Copy'')} that measures how much of the training data appearance leaks into the generated data. We use the SSCD description score \cite{pizzi2022self} to detect image copying between the training videos and the generated videos. 
The copying score is the maximum SSCD score between generated and training frames, with the percentage of samples above a manually set threshold of 0.25, indicating significant copying of the person or background.

\paragraphcustom{Ablation study.} 
We perform an ablation study and report motion accuracy (``Accuracy'') and copying score (``Copy'') in Table \ref{tab:ablations}.
\begin{table*}[!t]
 \caption{{\bf Quantitative results of the ablation study.} Each table examines the design choices of our method. We report the motion recognition accuracy (``Accuracy'') obtained with a pre-trained classifier for gesture recognition. The copying score (``Copy'') is the percentage of generated videos with a detection score above a set threshold.}
\footnotesize
    \centering
\hspace{\fill}
\begin{subtable}[ht]{0.2\textwidth}
\centering
\caption{Spatial layers}

\resizebox{1.2\textwidth}{!}{%
\begin{tabular}{c|c|c} %
Spatial layers & Accuracy $\uparrow$ & Copy $\downarrow$ \\ \cline{1-3} %
\hline
None & 62.3 & {\bf 5.9} \\
All & 68.7 & 17.0 \\ 
K,V (Ours) & {\bf 70.6} & 8.7 \\ 

\end{tabular}
}
\label{tab:spatial_layers}

\end{subtable}%
\hspace{\fill}
\hspace{\fill}
\begin{subtable}[ht]{0.2\textwidth}
\centering
\caption{Temporal layers}

\resizebox{1.2\textwidth}{!}{%
\begin{tabular}{c|c|c} %
Trans.\ layers & Accuracy $\uparrow$ & Copy $\downarrow$ \\ \cline{1-3} %
\hline
None & 22.7 & 1.4 \\ 
All (Ours) & {\bf 70.6} & 8.7 \\ 
\end{tabular}
}
\label{tab:temporal_layers}

\end{subtable}%
\hspace{\fill}
\hspace{\fill}
\begin{subtable}[ht]{0.2\textwidth}
\centering
\caption{Text Token}

\resizebox{1.2\textwidth}{!}{%
\begin{tabular}{c|c|c} %
Text Token & Accuracy $\uparrow$ & Copy $\downarrow$ \\ \cline{1-3} %
\hline
\checkmark & {\bf 75.5} & 22.7  \\ 
\xmark \ (Ours) & 70.6 & {\bf 8.7}  \\ 
\end{tabular}
}
\label{tab:text_token}

\end{subtable}%
\hspace{\fill}

\hspace{\fill}
\begin{subtable}[ht]{0.2\textwidth}
\caption{Sampling \\ Strategy}

\centering
\resizebox{1.2\textwidth}{!}{%
\begin{tabular}{l|c|c} %
Sampling & Accuracy $\uparrow$ & Copy $\downarrow$ \\ \cline{1-3} %
\hline
Uniform & 66.9 & 15.4 \\ 
Coarse-noise (Ours) & {\bf 70.6} & {\bf 8.7} \\ 
\end{tabular}}
\label{tab:sampling}

\end{subtable}%
\hspace{\fill}
\hspace{\fill}
\begin{subtable}[ht]{0.2\textwidth}
\caption{Regularization}

\centering
\resizebox{1.2\textwidth}{!}{%
\begin{tabular}{l|c|c} %
Prior  & Accuracy $\uparrow$ & Copy $\downarrow$ \\ \cline{1-3} %
\hline
None & 43.9 & {\bf 1.2} \\ 
Synthetic & 48.3 & 3.7 \\
Real (WebVid10M) & 66.7 & 10.9 \\
Real (Jester) & {\bf 70.6} & 8.7 \\
\end{tabular}
}
\label{tab:regularization}

\end{subtable}%
\hspace{\fill}
\hspace{\fill}
\begin{subtable}[ht]{0.2\textwidth}
\centering
\caption{Fine-tuning \\ Strategy}

\resizebox{1.2\textwidth}{!}{%
\begin{tabular}{c|c|c|c} %
Tuning Method & Layers & Accuracy $\uparrow$ & Copy $\downarrow$ \\ \cline{1-3} %
\hline
LoRA & All & 17.4 & \textbf{0.5} \\
LoRA & Ours & 10.6 & 3.3 \\ 
Full & All & 68.7 & 17.0 \\
Full & Ours & {\bf 70.6} & 8.7 \\

\end{tabular}
}
\label{tab:lora}

\end{subtable}%
\hspace{\fill}
    \label{tab:ablations}
\end{table*}

First, we study the choice of the optimized parameters. As in the case of the image customization methods, it is clear that optimizing the spatial layers of the model is not enough to learn a novel motion pattern, as shown in Table \ref{tab:temporal_layers}. Training the key and value projections in the cross-attention layers of the spatial layers achieves a good balance in terms of accuracy and copying score, as shown Table \ref{tab:spatial_layers}. We observe a two-fold reduction in the copying score when training only those parameters and an 8\% improvement in motion accuracy compared to not training any spatial parameters. 

Next, in Table \ref{tab:text_token}, we consider the role of optimizing the text token embedding. We notice that when jointly training the text token embedding with the model's parameters, the motion accuracy is higher than when training only the model parameters. However, it also leads to nearly three times as much memorization of the training video appearance, as evidenced by the significantly increased Copy score. We hypothesize that this indicates that the text token embedding is learning something related to the appearance of the training videos. 
To test our sampling strategy we compare a model trained with our motion pattern sampling strategy (``Coarse-noise'') to a model that simply samples diffusion time steps uniformly (``Uniform''). As shown in Table \ref{tab:sampling}, our sampling strategy improves the motion accuracy and reduces the appearance copying. Following our intuition regarding learning motion patterns rather than fine-grained details, our model is less prone to replicating the training examples.

We study the effect of the regularization prior in Table \ref{tab:regularization}. We train four models with the same design choice as our method, yet we choose the regularization set to be either: (i) empty, (ii) a synthetic set of videos generated from the original text-to-video model, or (iii) real videos from the WebVid10m dataset (iv) containing real videos that contain similar visual scenes. We observe a significant drop in performance both without using a prior as well as using a synthetic prior. Using real data is beneficial for fine-tuning. We observe that using motions and visual scenes similar to the target motion is optimal. However, should this support set be unavailable, off-the-shelf dataset is still better than no regularization or synthetic video. Overall, our method achieves a good balance between high motion accuracy and low copying score. Lastly, we compare full parameter fine-tuning to low-rank adaptation in Table \ref{tab:lora}. We test two variants; adding LoRA adapters to both spatial and temporal transformers (first row), and to the same layers as our NewMove (second row). In our experiments we use rank 32 and generate videos with LoRA scale 1, we observe that full parameter fine-tuning achieves higher motion accuracy. Qualitatively, we observe that LoRA models perform relatively well on coarser gestures like \textit{Drumming Fingers} but struggle to learn fine-grained motions like \textit{Sliding Two Fingers Up}.

\paragraphcustom{Quantitative evaluation of motion fidelity.}
We quantitatively evaluate our approach by computing metrics corresponding to the quality of the generated motion and the overall fidelity of the generated video with respect to the input text prompt. We compare to baseline methods that have explicitly learned the motion pattern from a few examples. 
We follow the evaluation protocol as defined in Section \ref{sec:quantitative}. We report the motion accuracy and text alignment metrics across different methods in Table \ref{tab:soa} and show qualitative comparison (Figure \ref{fig:sota}). 
\setlength{\intextsep}{1pt}  %
\begin{wraptable}[10]{r}{0.6\linewidth}
\caption{{\bf Quantitative comparison with baseline methods}}
\centering
\resizebox{\linewidth}{!}{

\centering
\begin{tabular}{lcc}
\toprule
 & Motion accuracy $\uparrow$& Text Alignment $\uparrow$ 
 \\
 
\midrule
Textual Inversion \cite{gal2022image} & 0.3 & 0.2733 \\
Custom Diffusion \cite{kumari2023multi} & 10.5 & 0.2788 \\
Dreambooth \cite{ruiz2023dreambooth} & 28.4 & 0.2796 \\
Tune-a-Video \cite{wu2023tune} & 18.9 & {\bf 0.2818} \\
MotionDirector \cite{zhao2023motiondirector} & 24.7 & 0.2786  \\

Ours & {\bf 70.6} & {\bf 0.2818} \\
\bottomrule
\end{tabular}

}
\label{tab:soa}
\end{wraptable}
We observe that Textual Inversion completely fails to learn the motion, this failure is potentially because the text encoder has been trained only on image-text pairs, and the embedding fails to generalize to the video domain. Additionally, the text embedding is injected only into spatial layers in the cross-attention layers, and not the temporal ones. Alternatively, because we are learning an unseen motion pattern, which is more complex than a static object, the embedding does not have enough parameters to learn it. Dreambooth and Custom Diffusion learn to adhere to the spatial structure of the training videos and some hand manipulation. However they fail to accurately reproduce the motion and produce a temporally smooth video. Our approach yields over a 2$\times$ improvement in motion accuracy and more faithfully adheres to the text prompt. We also compare with the concurrent MotionDirector method \cite{zhao2023motiondirector} and find that our approach outperforms it in motion accuracy and text alignment. However, with longer training, the baseline achieves similar performance (see Appendix for details and qualitative comparison).

\section{Conclusion}
We present an approach for motion customization in text-to-video diffusion models. Our method can learn a new motion pattern from a set of few exemplar videos of different subjects performing the same motion. We conduct a thorough ablation study that identifies the key components of the method and evaluate them with respect to motion accuracy. We demonstrate qualitative results of our method and evaluate our method with a user study, specifically in scenarios where our method generalizes the motion to unseen actors, multiple people performing the motion, and different viewpoints.

\bibliographystyle{splncs04}
\bibliography{main}

\end{document}